# Improving The Diagnosis of Thyroid Cancer by Machine Learning and Clinical Data

Nan Miles Xi [a], Lin Wang [b], and Chuanjia Yang [c,*]

[a] Department of Mathematics and Statistics, Loyola University Chicago, Chicago, IL 60660, USA

[b] Department of Statistics, The George Washington University, DC, 20052, USA

[c] Department of General Surgery, Shengjing Hospital of China Medical University, Shenyang, Liaoning 110004, China

[*] Correspondence: Chuanjia Yang (cjyang@cmu.edu.cn)

## Abstract

Thyroid cancer is a common endocrine carcinoma that occurs in the thyroid gland. Much effort has been invested in improving its diagnosis, and thyroidectomy remains the primary treatment method. A successful operation without unnecessary side injuries relies on an accurate preoperative diagnosis. Current human assessment of thyroid nodule malignancy is prone to errors and may not guarantee an accurate preoperative diagnosis. This study proposed a machine framework to predict thyroid nodule malignancy based on a novel clinical dataset we collected. The 10-fold cross-validation, bootstrap analysis, and permutation predictor importance were applied to estimate and interpret the model performance under uncertainty. The comparison between model prediction and expert assessment shows the advantage of our framework over human judgment in predicting thyroid nodule malignancy. Our method is accurate, interpretable, and thus useable as additional evidence in the preoperative diagnosis for thyroid cancer.

# Introduction

Thyroid cancer is the most frequent endocrine malignancy and represents about 2.5% of all new cancer cases in the United States [1]. According to NIH's Surveillance, Epidemiology, and End Results Program (SEER), the occurrence of thyroid cancer has increased by 5.5% annually from 2005 to 2015 [2]. In the United States, the annual incidence rate stands at 14.1 per 100,000 between 2014 and 2018. The annual death rate is 0.5 per 100,000 between 2015 and 2019. The American Cancer Society estimated that there would be 43,800 new cases and 2,230 deaths caused by thyroid cancer in 2022 [3]. Among thyroid cancers, 96% originate from follicular cells, and of these, 99% are differentiated thyroid cancer (DTC) [4]. The treatment methods for DTCs, especially papillary thyroid cancer (PTC), mainly include surgery, TSH suppressive therapy with levothyroxine, and radioactive iodine remnant ablation [5]. While individualized treatment depends on the nature of the lesion, surgical operation remains the primary tool at present. One focus of the surgical operation is to distinguish between benign and malignant thyroid nodules. An accurate preoperative diagnosis is conducive to a smooth operation, avoids unnecessary side injuries, and reduces the risk of post-operative recurrence [6]. It also helps the selection of comprehensive post-surgery treatment to extend the survival period. Therefore, it is crucial to make accurate diagnosis and prediction based on thyroid ultrasound, blood test, and other basic clinical information.

To date, the diagnosis of malignant nodules largely relies on the clinical experience of surgeons and radiologists [7]. In many cases, human judgment is time-consuming and prone to error. Accurate and explainable predictive models are urgently needed to help medical decisions and reduce labor work. Previous studies have built statistical models to predict the occurrence of malignant nodules based on various datasets [8–11]. Those models mainly utilized descriptive statistics or logistic regression, which ignored the nonlinear relationship among clinical and demographical variables. Other machine learning-based models only provided the point estimation of the model performance without considering the uncertainty in the model prediction [12]. More seriously, no study has compared the diagnostic accuracy between model prediction and expert assessment. The lack of fair comparisons makes it difficult to evaluate the advantage of using predictive models to assist the diagnosis of malignant nodules.

In this paper, we proposed a comprehensive machine learning framework to accurately predict nodule malignancy. We collected a novel clinical dataset containing 724 patients with 1232 nodules. We trained six cutting-edge machine learning models on this dataset and estimated their unbiased prediction performance by 10-fold cross-validation. The uncertainty of model performance was further quantified by bootstrap analysis. We identified the important variables through the analysis of normalized permutation predictor importance. Finally, we compared the model performance to expert assessment and demonstrated the advantage of the machine learning model over human judgment. In general, the proposed machine learning models exhibited high prediction accuracy and diverse capacity to identify benign or malignant nodules.

This result is consistent under both point estimation and model uncertainty analysis. Many of the identified important variables confirm similar findings in previous studies. The best-performed models outperformed the expert assessment by a large margin on the same dataset, which indicates the benefits of using machine learning models to improve the preoperative diagnosis of nodule malignancy.

## Data Collection and Preprocessing

The dataset used in this study was collected from 724 patients who were admitted to Shengjing Hospital of China Median University between 2010 and 2012. All patients underwent thyroidectomy, and their nodule malignancy, demographic information, ultrasound features, and blood test results were recorded in the datasets. For each patient, we observed one or multiple nodules located in three areas, i.e., left lobe, right lobe, and isthmus. If the patient had multiple nodules in one area, we kept the largest one in the dataset. After removing missing values, there are 1232 nodules and 19 variables in the dataset. The descriptive statistics of those nodules and variables are described in Table 1.

The average age of patients is 46.61, with a range of 13-82. There are 200 male-patient-affiliated nodules (16.23%) and 1032 female-patient-affiliated nodules (83.77%). We recorded five thyroid function tests, including free triiodothyronine (FT3), free thyroxine (FT4), thyroid-stimulating hormone (TSH), thyroid peroxidase antibodies (TPO), and thyroglobulin antibodies (TgAb). All thyroids are categorized into even (89.12%) and uneven (10.88%) based on their echogenicity. There are 11 variables that describe the characteristics of nodules: 1) size is defined as the maximum between the length and width of each nodule; 2) location is left lobe, right lobe, or isthmus for each nodule; 3) multifocality indicates if there are multiple nodules identified in one location; 4) shape describes the nodule's regularity; 5) margin characterizes if the nodule has a clear or unclear margin; 6) calcification suggests the existence of nodule calcification; 7) echogenicity describes the nodule's ability to bounce echoes in ultrasound; 8) blood flow is defined as normal or enriched for each nodule; 9) nodule's composition is categorized into cystic, mixed, and solid; 10) laterality demonstrates if the nodule has counterparts in other locations of the same patient; 11) malignancy is determined by examining the nodule specimen after thyroidectomy.

## Methods

We utilized gradient boosting machine (GBM) [13], logistic regression, linear discriminant analysis (LDA) [14], support vector machine (SVM) with radial or linear kernel [15], and random forest [16] to train six machine learning models to predict the nodule malignancy based on the dataset described in the last section. In the predictive models, the malignancy was treated as a response, and the other 18 variables were predictors. We conducted 10-fold cross-validation to obtain an unbiased estimation of prediction accuracy [17]. First, we randomly split the patients into 10 groups. Second, one patient group was selected, and its affiliated nodules were used to generate the test set. Nodules of the other nine patient groups were treated as the training set. Third, we trained the machine learning models on the training set and then predicted the nodule malignancy in the test set. Finally, we performed the same process until every patient group and their affiliated nodules were predicted by the machine learning models. We repeated the 10-fold cross-validation by 10 times to reduce the variability introduced by random splitting.

We compared the model prediction with the true nodule malignancy to evaluate the model performance. For each model, we calculated the accuracy, area under the receiver operating characteristic (AUROC), sensitivity, specificity, and precision. The accuracy is the proportion of correct predictions among all nodules in the dataset. The AUROC measures the overall diagnostic ability of a binary predictive model as its discrimination threshold is varied [18]. The sensitivity represents the proportion of malignant nodules that are correctly predicted as malignant. The specificity represents the proportion of benign nodules that are correctly predicted as benign. The precision is defined as the proportion of true malignant nodules among those predicted as malignant. These five measurements together provide a comprehensive summary of the diagnostic capacity of predictive models. We implemented the model training and evaluation by the R programming language [19].

## Results

**Overall model performance.** Table 2 summarizes the accuracy, AUROC, sensitivity, specificity, and precision for six machine learning models measured by 10-fold cross-validation. All the measurements are averaged across 10 repetitions described in the Methods section. Among all models, random forest achieves the highest prediction accuracy (0.7931) and AUROC (0.8541), which indicates a good overall capacity to differentiate between benign and malignant nodules. The GBM model outperforms others in terms of sensitivity (0.8750). The high sensitivity of the GBM model shows its strongness to find malignant nodules. On the other hand, logistic regression has advantages on specificity (0.6806) and precision (0.8384). Unlike the GBM model, logistic regression is more capable of identifying benign nodules. Also, the high precision of logistic regression means that among its predicted malignant nodules, a large number of them are truly malignant. Overall, machine learning models exhibit mixed performance in predicting nodule malignancy. There is no single model dominating others on all five measurements. We can choose different models according to the specific requirement in diagnosing malignant nodules.

**Model uncertainty measurement.** The five measurements in Table 2 are point estimations of the model performance. To further understand the uncertainty of the model prediction, we conducted a bootstrap to construct the empirical distributions of the five model performance measurements [17]. In each step of the 10-fold cross-validation, we resampled with replacement from the original training set to generate a bootstrap training set. Then we trained machine learning models on this bootstrap training set and evaluated its prediction performance on the test set. We repeated this resampling 1000 times and followed the previous model training and evaluation process to obtain the empirical distributions of prediction accuracy, AUROC, sensitivity, specificity, and precision.

Figure 1 and Table 3 compare those five empirical distributions and their summary statistics. The six machine learning models show a similar asymptotical performance ranking compared with their point estimation in Table 2. The SVM with linear kernel has the highest median prediction accuracy, followed by random forest and SVM with radials kernel. The random forest and GBM outperform other models on AUROC, and the GBM also shows advantages in terms of sensitivity. The logistic regression is the best-performed model measured by precision and specificity. Although the two SVM models perform relatively well in general, they introduce some low accuracy in all five measurements, reflected by the small outliers in Figure 1. We observe that the performance difference among the six models is relatively small on the accuracy and AUROC, the two general accuracy measurements. However, the gaps are more significant if measured by precision, sensitivity, and precision, indicating diverse model behavior in the differentiation of malignant and benign nodules.

**Variable importance analysis.** We conducted a variable importance analysis to examine the different effects of nodule characteristics on the model performance. We utilized the permutation predictor importance to measure the contribution of each variable to the model prediction [16]. The permutation predictor importance of one variable is defined as the decrease of the AUROC when that variable's value is randomly shuffled. Since the random shuffling breaks the relationship between the variables (characteristics of patients and nodules) and response (nodule malignancy), any decrease of AUROC indicates the model's dependency on that shuffled variable. Using permutation predictor importance has three advantages. First, its calculation does not rely on the specific model form. Second, the predictive model only needs to be trained once. Third, the random shuffling can be repeated multiple times to reduce the variability in the calculation.

For each variable, we averaged its permutation predictor importance across all six models. Then we normalized each average by their maximum among all variables to obtain the final normalized permutation predictor importance. Figure 2 shows the top 10 variables ranked by their normalized permutation predictor importance. Calcification has the most substantial impact on the prediction of nodule malignancy, followed by laterality, blood flow, and location. The composition and size have similar predictor importance, less than half of the top variable calcification. The shape of nodules is the last-tier important variable, with only 20% importance as the calcification. Other variables have significantly less impact on the model performance.

In addition to identifying the important variables, we further explored how they would impact the prediction of nodule malignancy. Figure 3 shows the percentage of malignant nodules corresponding to each value of top-six important variables. A large percentage (close to 100%) indicates that the specific value of that variable is an indicator of malignant nodules. A small percentage (close to 0%) indicates that the specific value of the variable is an indicator of benign nodules. A close-to-50% percentage shows no strong indication of the nodule malignancy. We find that the presence of calcification, unilateral, enriched blood flow, left- or right-located, solid composition, and larger size (greater than 0.8 cm) are strong indicators of malignant nodules. On the other hand, a cystic nodule located at the isthmus is more likely benign. It should be mentioned that Figures 2 and 3 emphasize the marginal effects of variables on the model prediction. The non-top important variables may also impact the model performance through interactions with other variables.

**Comparison between model prediction and expert assessment**. One objective of building machine learning models is to assist the diagnosis of malignant thyroid nodules before surgery. To evaluate how well the proposed models fulfilled this objective, we performed a comparative analysis between model prediction and expert assessment. First, we removed the true nodule malignancy from the original dataset. Second, we provided the dataset to a surgeon with 16 years of experience specializing in thyroid cancer. The surgeon was asked to assess the malignancy of all 1232 nodules in the dataset. Third, we compared the prediction results from the surgeon with the ones from the machine learning models. It is worth noting that the surgeon and models have the same access to the variables in Table 1. Therefore, the comparison is fair. We will refer to the surgeon as the expert in the following text.

Table 4 compares the five prediction measurements between the expert assessment and random forest, the best-performed model in Table 2. Unlike the model prediction, the expert assessment has no probability of being benign or malignant output. Therefore, the AUROC cannot be used in this comparison. Instead, we use F1 score, the harmonic mean of the precision and sensitivity, to evaluate the overall diagnosis [20]. The F1 score measures the model's overall capacity to identify malignant nodules. We found that the random forest outperformed expert assessment on the accuracy, F1 score, and sensitivity, with leading margins as 12%, 14%, and 27%, respectively. The measurements on which the expert assessment shows advantage are specificity and precision, with leading margins as 17% and 4%, respectively.

The comparison results in Table 4 demonstrate the model's superior predictive performance over the expert assessment. In summary, the random forest is 1) more accurate than the expert in general (higher accuracy); 2) more capable of finding malignant nodules from the dataset (higher sensitivity and F1 score). On the other hand, the expert tends to find benign nodules more efficiently (higher specificity). Even though the expert assessment is slightly more accurate among predicted malignant nodules (higher precision), the higher F1 score of the random forest indicates its overall stronger capacity in identifying malignant nodules.

To understand the predictive behavior of the expert and random forest, we compared their confusion matrices in Table 5. Among the correct predictions (diagonal elements), the random forest found more malignant nodules than the expert (714 vs. 487). In contrast, the expert identified more benign nodules than the random forest (342 vs. 272). This result echoes the findings in Table 4, where random forest shows high sensitivity (true malignant rate) and expert shows high specificity (true benign rate). Among the wrong predictions (off-diagonal elements), the random forest overestimated more nodules' malignancy than the expert (141 vs. 71). However, the expert underestimated more nodules' malignancy than the random forest (331 vs. 105). This comparison implies an opposite predictive behavior – the expert is more conservative in predicting nodules as malignant, while the random forest is more aggressive in the prediction. The errors caused by aggressive prediction are less than those caused by conservative prediction, making the random forest more accurate than the expert assessment overall.

## Discussion

In this study, we utilized machine learning methods to assist the diagnosis of malignant thyroid nodules. We collected a real dataset with 724 patients' demographic and clinical information. 1232 nodules from those patients are included, and their malignancy is confirmed by thyroidectomy. Based on this dataset, we built six machine learning models to predict the malignancy of thyroid nodules. We used five measurements to provide a comprehensive evaluation of the model performance. Although no single model outperforms others among all measurements, the decision-tree-based nonlinear models, i.e., random forest and GBM, exhibit better overall diagnostic accuracy (measured by accuracy and AUROC) and the capacity to identify malignant nodules (measured by sensitivity). Similar model performance is observed in both point estimation and uncertainty measurement (Tables 2 and 3). The linear predictive model, logistic regression, is good at finding benign nodules (measured by specificity). Consequently, random forest and GBM are more suitable for early cancer screening, but at the expense of false malignant diagnosis. Interestingly, the logistic regression made most correct predictions among predicted malignant nodules, which is shown by the highest precision in point estimation and uncertainty measurement. Thus, the logistic regression can be ensembled with random forest or GMB to improve the model's diagnostic capacity.

Overall, the machine models exhibit satisfactory prediction performance. The average accuracy and AUROC of the five models are 0.78 and 0.84, respectively. In practice, an AUROC greater than 0.8 indicates excellent discrimination between binary outcomes [18]. One encouraging result of our study is the superior model performance over the expert assessment. The best-performed model, random forest, beat the expert assessment by 12% on accuracy, and 14% on F1 score, the two general measurements. One interpretation of better prediction by machine learning models is that they are able to capture the complex nonlinear relationships among different variables. Such relationships are implicitly contained in the dataset and difficult for humans to identify. The models are also more aggressive in predicting nodules as malignant. As a result, the machine learning model is valuable for diagnosing thyroid cancer.

Our variable importance analysis identified key variables in the diagnosis of malignant thyroid nodules (Figures 2 and 3). Those variables are consistent with previous findings in clinical and modeling studies. For example, it has been recognized that the existence of calcification, cystic composition, and large nodule size are strong indicators for malignancy [8–12]. Such consensus is confirmed in our analysis. Regarding other important variables, there are some debates about the role of blood flow in the diagnosis of malignant nodules [21]. Our study suggests that the nodules with enriched blood flow are more likely to be malignant. One of the important variables, laterality, was largely ignored in previous studies. We find that the unilateral nodules have a high possibility of becoming malignant. We suspect the potential reason is that in the ultrasound examination, if nodules were observed in multiple locations (left lobe, right lobe, or isthmus), only those with a high likelihood of being malignant were recorded. Therefore, there tend to be more unilateral malignant nodules in the dataset. The nodule location is also recognized by the model as an important variable. However, being located at the isthmus reduces the possibility of malignancy, which contradicts previous literature [22]. More data from larger patient cohorts would be necessary to investigate the true impact of nodule location further.

Several topics are worth further exploration. First, the current dataset contains 724 patients with 1232 nodules. Although the sample size is not small, collecting more data will increase the diversity of patients and nodules. The model trained on a more extensive and diverse dataset would generalize better to new patients when deployed in real-world scenarios. Second, the ultrasound-related variables in our dataset are extracted by sonographers from the original ultrasonography. Such extraction may omit important features only detectible in the raw images. A deep convolutional neural network can be applied to catch those features directly from the ultrasonography and improve the model performance [23]. Third, we can encompass novel techniques beyond traditional ultrasound and blood test into the modeling process. For example, single-cell RNA-sequencing (scRNA-seq), a cutting-edge sequencing technology revealing the genome-wide gene expression at single-cell levels, can be applied to evaluate and compare the transcriptomes of thyroid nodules [24–26]. The gene expression dynamic associated with nodule malignancy will potentially improve the model performance, similar to the progress made by scRNA-seq in other cancer diagnoses and precision medicine [27,28]. Finally, the machine learning framework used in this study can evaluate the quality of clinical data for thyroid cancer diagnosis [29]. A machine learning model can be trained on datasets collected from different studies. A high-quality dataset is expected to contain enough information for models accurately predicting nodule malignancy. Therefore, the model prediction accuracy will serve as a proxy for data quality of different datasets.

**Data and Code Availability**

The data used in this study is available at Zenodo repository:

https://zenodo.org/record/6387087#.Yj_KeeqZM4c

The source code that implemented the result in this study is available at GitHub repository:

https://github.com/xnnba1984/Thyroid-Cancer

# Tables

| Patient Characteristics | Value | Percentage |
|---|---|---|
| Age (years) | | |
| Mean ± SD | 46.61 ± 12.44 | |
| Range | 13 – 82 | |
| Gender | | |
| Male | 200 | 16.23% |
| Female | 1032 | 83.77% |
| **Test (Median ± IQR)** | | |
| FT3 | 4.35 ± 0.82 | |
| FT4 | 14.51 ± 2.56 | |
| TSH | 1.46 ± 1.63 | |
| TPO | 0.63 ± 5.37 | |
| TgAb | 2.69 ± 11.88 | |
| **Thyroid Characteristics** | | |
| Echogenicity | | |
| Even | 1098 | 89.12% |
| Uneven | 134 | 10.88% |
| **Nodule Characteristics** | | |
| Size (Mean ± SD, cm) | 1.73 ± 1.31 | |
| Location | | |
| Right | 584 | 47.40% |
| Left | 548 | 44.48% |
| Isthmus | 100 | 8.22% |
| Multifocality | | |
| Unifocal | 664 | 53.90% |
| Multifocal | 568 | 46.10% |
| Shape | | |
| Regular | 977 | 79.30% |
| Irregular | 255 | 20.70% |
| Margin | | |
| Clear | 406 | 32.95% |
| Unclear | 826 | 67.05% |
| Calcification | | |
| Absent | 740 | 60.06% |
| Present | 492 | 39.94% |
| Echogenicity | | |
| None | 16 | 1.30% |
| Isoechoic | 15 | 1.21% |
| Medium-echogenic | 144 | 11.69% |
| Hyperechogenic | 7 | 0.57% |
| Hypoechogenic | 1050 | 85.23% |
| Blood Flow | | |
| Normal | 786 | 63.80% |
| Enriched | 446 | 36.20% |
| Composition | | |
| Cystic | 30 | 2.44% |
| Mixed | 97 | 7.87% |
| Solid | 1105 | 89.69% |
| Laterality | | |
| Unilateral | 286 | 23.21% |
| Multilateral | 946 | 76.79% |
| Malignancy | | |
| Benign | 413 | 33.52% |
| Malignant | 819 | 66.48% |

**Table 1**. **The characteristics of patients, blood tests, thyroids, and nodules in the dataset used in this study.**

| Model | Accuracy | AUROC | Sensitivity | Specificity | Precision |
|---|---|---|---|---|---|
| GBM | 0.7741 | 0.8497 | <u>0.8750</u> | 0.5741 | 0.8029 |
| Logistic | 0.7834 | 0.8422 | 0.8352 | <u>0.6806</u> | <u>0.8384</u> |
| LDA | 0.7790 | 0.8394 | 0.8452 | 0.6477 | 0.8263 |
| SVM (Radial) | 0.7688 | 0.8237 | 0.8435 | 0.6206 | 0.8149 |
| SVM (Linear) | 0.7661 | 0.8200 | 0.8322 | 0.6349 | 0.8186 |
| Random Forest | <u>0.7931</u> | <u>0.8541</u> | 0.8629 | 0.6547 | 0.8321 |

**Table 2**. **The model prediction performance measured by five measurements.** Each measurement was calculated by 10-fold cross-validation and then averaged across 10 repetitions. The highest values among the six models are underscored.

| Model | Measurement | Accuracy | AUROC | Sensitivity | Specificity | Precision |
|---|---|---|---|---|---|---|
| GBM | 95% CI | (0.7605, 0.7833) | (0.8359, 0.8469) | (0.8569, 0.8901) | (0.5327, 0.6029) | (0.7892, 0.8126) |
| | Mean | 0.7722 | 0.8432 | <u>0.8753</u> | 0.5678 | 0.8007 |
| Logistic | 95% CI | (0.7646, 0.7881) | (0.8238, 0.8398) | (0.8205, 0.8535) | (0.6223, 0.6901) | (0.8149, 0.8417) |
| | Mean | 0.7770 | 0.8323 | 0.8370 | <u>0.6579</u> | <u>0.8292</u> |
| LDA | 95% CI | (0.7634, 0.7857) | (0.8219, 0.8375) | (0.8296, 0.8670) | (0.5906, 0.6659) | (0.8055, 0.8328) |
| | Mean | 0.7737 | 0.8308 | 0.8473 | 0.6280 | 0.8188 |
| SVM (Radial) | 95% CI | (0.7240, 0.7930) | (0.7711, 0.8470) | (0.7851, 0.8718) | (0.5956, 0.6489) | (0.7944, 0.8292) |
| | Mean | 0.7798 | 0.8378 | 0.8560 | 0.6288 | 0.8205 |
| SVM (Linear) | 95% CI | (0.7224, 0.7898) | (0.7694, 0.8439) | (0.7790, 0.8608) | (0.6077, 0.6586) | (0.7975, 0.8323) |
| | Mean | 0.7779 | 0.8343 | 0.8466 | 0.6417 | 0.8240 |
| Random Forest | 95% CI | (0.7670, 0.7930) | (0.8363, 0.8523) | (0.8449, 0.8743) | (0.5860, 0.6538) | (0.8060, 0.8310) |
| | Mean | <u>0.7801</u> | <u>0.8443</u> | 0.8605 | 0.6208 | 0.8182 |

**Table 3**. **The summary statistics of model performance calculated by bootstrap.** The empirical 95% confidence intervals and means of five measurements were calculated for each model. The highest mean values among the six models are underscored.

| Method | Accuracy | F1 | Sensitivity | Specificity | Precision |
|---|---|---|---|---|---|
| Expert Assessment | 0.6734 | 0.7078 | 0.5954 | <u>0.8281</u> | <u>0.8728</u> |
| Random Forest | <u>0.7931</u> | <u>0.8472</u> | <u>0.8629</u> | 0.6547 | 0.8321 |

**Table 4**. **The comparison of five prediction measurements between expert assessment and random forest.** The highest values between the two are underscored.

**Expert Assessment**

|  |  | Truth | |
|---|---|---|---|
|  |  | Benign | Malignant |
| **Prediction** | Benign | 342 | 331 |
|  | Malignant | 71 | 487 |

**Random Forest**

|  |  | Truth | |
|---|---|---|---|
|  |  | Benign | Malignant |
| **Prediction** | Benign | 272 | 105 |
|  | Malignant | 141 | 714 |

**Table 5**. **The confusion matrices of expert assessment and random forest.** The diagonal elements are the correct predictions. The off-diagonal elements are wrong predictions.

# Figures

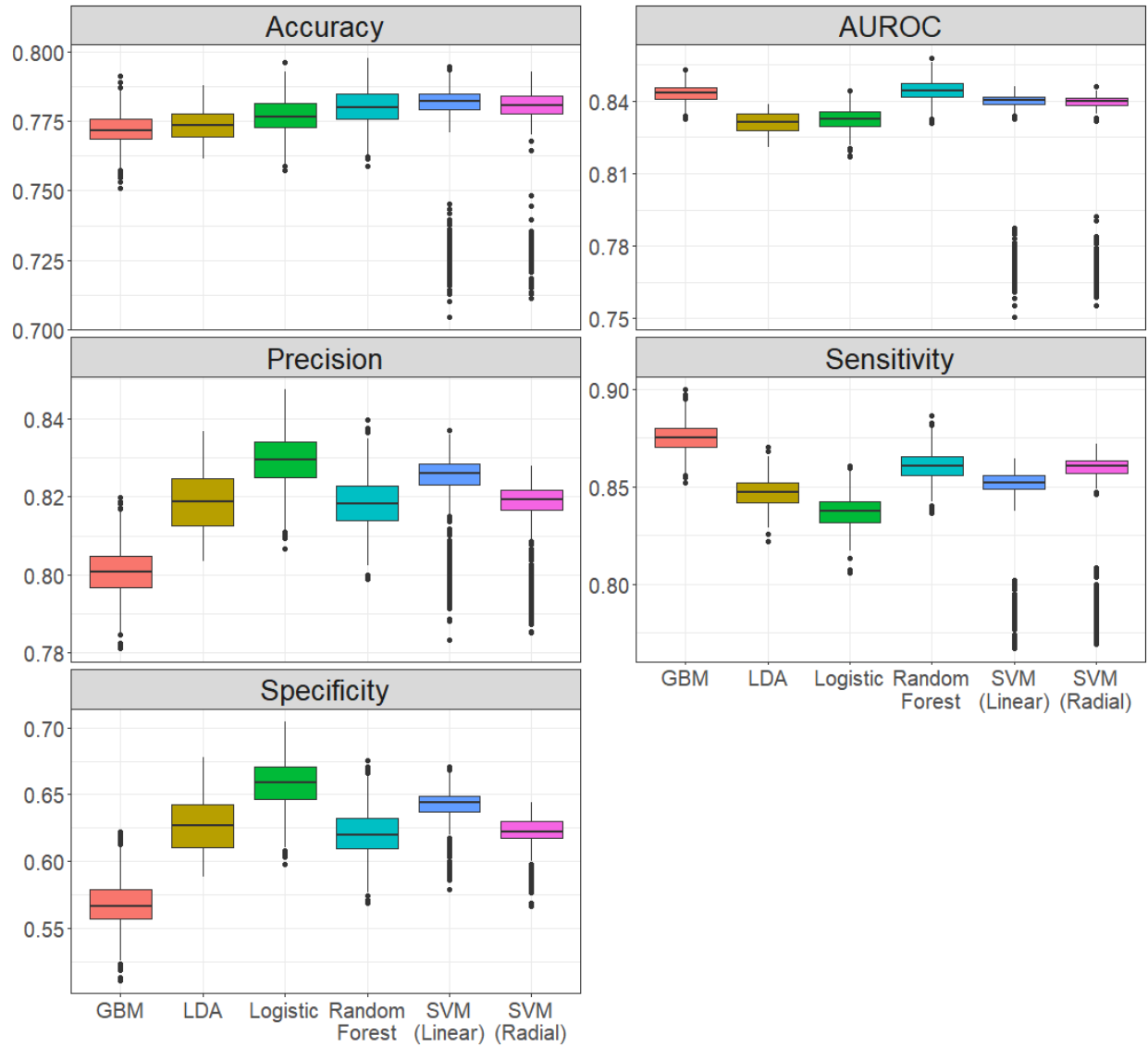

**Figure 1. The empirical distributions of model performance constructed by bootstrap.** Five performance measurements of six models were calculated on 1000 bootstrap samples.

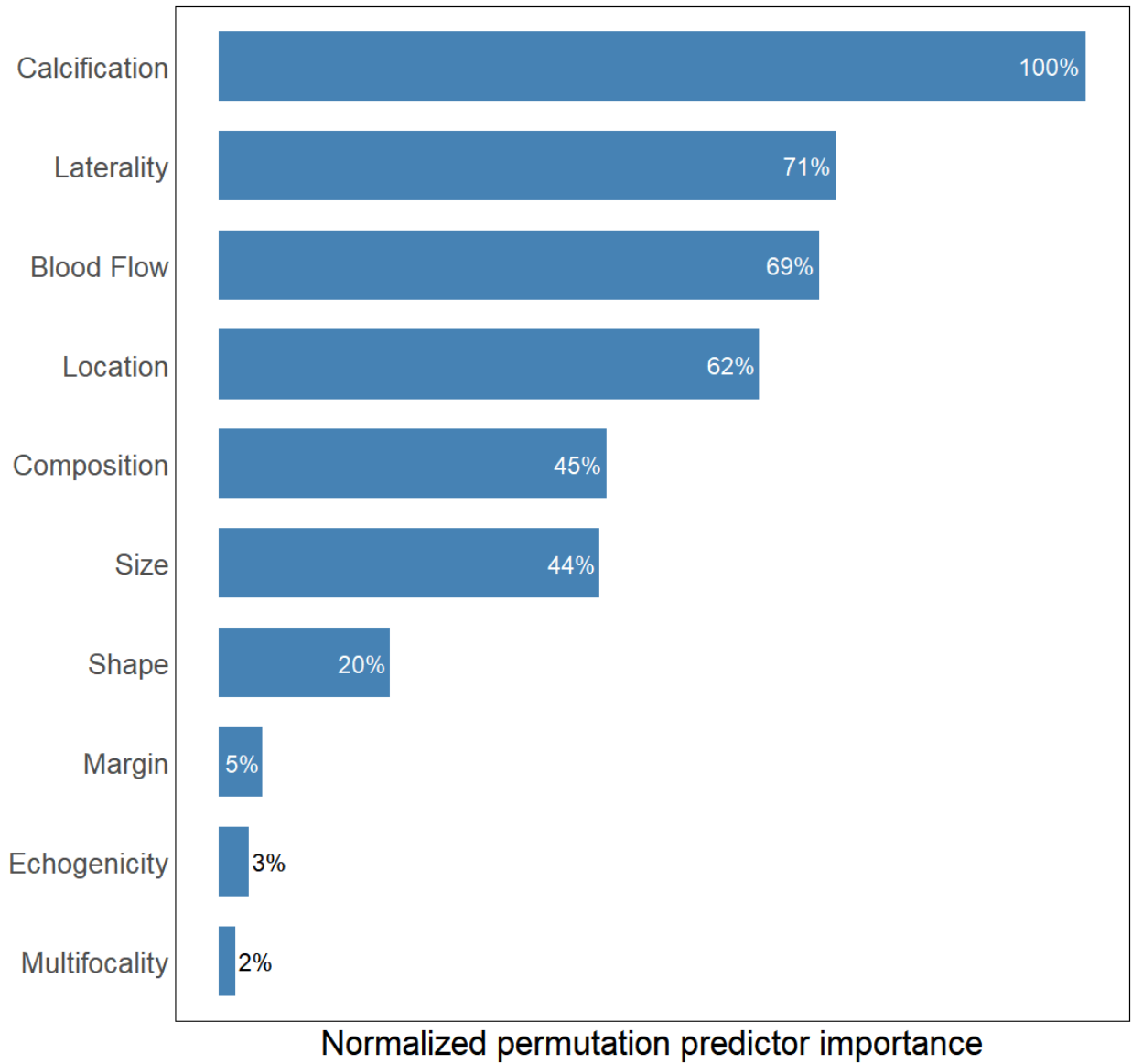

**Figure 2. The normalized permutation predictor importance for the top 10 variables**. Variables are sorted from high to low.

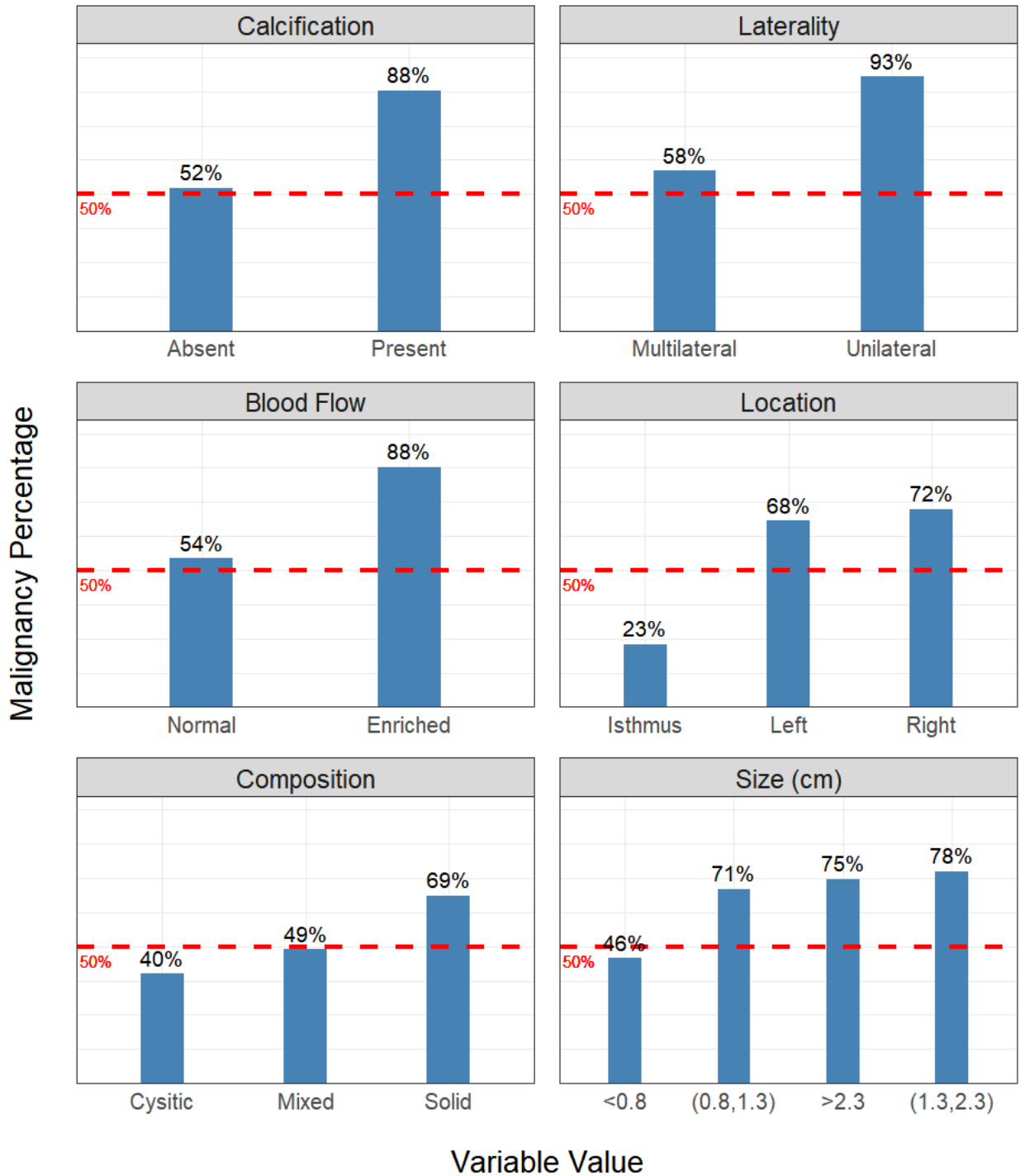

**Figure 3. The percentage of malignant nodules corresponding to each value of top-six important variables**. A large percentage shows that the specific value is an indicator of malignant nodules. A small percentage shows that the specific value is an indicator of benign nodules. A close-to-50% percentage shows no strong indication of the nodule malignancy.